\documentclass[11pt]{article}
\usepackage{setspace}
\usepackage{times}
\usepackage{algorithm}
\usepackage{algorithmic}
\usepackage{eucal}
\usepackage{amsfonts}
\usepackage{verbatim}

\def\sol{{\tt sol} }

\def\fun{{\tt fun} }
\def\img{{\tt img} }
\def\called{{\tt do} }

\def\C{{\cal C} }
\newcommand{\email}[1]{${\tt #1}$}
\newtheorem{proposition}{Proposition}
\newtheorem{example}{Example}

\title
{
	Constraint-based \mbox{analysis} of composite solvers\thanks
	{
		Financially supported
		by Centre Franco-Russe Liapunov (Project 06--98) and
		by European project COCONUT IST--2000--26063.
	}
}
\date{}

\author
{
	Evgueni Petrov\\
	Ecole des Mines de Nantes\\
	La Chantrerie, 4 rue Kastler, B.P.20722\\
	44307 Nantes Cedex 03 France\\
	\email{evgueni.petrov@emn.fr}
 \and
	\'Eric Monfroy\\
	Universit\'e de Nantes\\
	2 Houssini\`ere, 44300 Nantes Cedex 03 France\\
	\email{eric.monfroy@irin.univ-nantes.fr}
}

\begin{document}
\doublespace

\maketitle

\begin{abstract}
Cooperative constraint solving is an area of constraint programming
that studies the interaction between constraint solvers with the aim of discovering the interaction patterns that amplify the positive qualities of individual solvers.
Automatisation and formalisation of such studies is an important issue of cooperative constraint solving.

In this paper we present a constraint-based \mbox{analysis} of composite solvers that integrates reasoning about the individual solvers and the processed data.
The idea is to approximate this reasoning by resolution of set constraints on the finite sets representing the predicates that express all the necessary properties.
We illustrate application of our analysis to two important cooperation patterns: deterministic choice and loop.

\end{abstract}

\section{Introduction}
\label{intro}

Cooperative constraint solving is an area of constraint programming
that studies interaction between constraint solvers with the aim of discovering the interaction patterns that amplify the positive qualities of individual solvers.
Papers \cite{coop:jaffar-michaylov-stuckey-yap-acm-92,coop:beringer-backer-lp-elsevier-95,coop:marti-rueher-jait-95,vH-numerica,coop:hickey-lncs-00}
describe some examples of successful cooperative constraint solving systems.

At the present time, successful patterns of cooperation are detected with the sole help of
intuition, feelings, experience, and experiments with software engineering tools supporting cooperative constraint solving
such as \cite{ECLIPSE00,PROLOGIV,MOZART,coop:monfroy-frocos-00}.
One cannot make such experiments systematic without a mathematical framework for analysis of cooperative constraint solving systems (``composite solvers'').

An important practical application for such a framework is integration and development of domain-specific software.
The expected applications for the analysis are
detection of inconsistencies in the specifications of individual software packages and 
construction of the specifications for the packages needed in order to meet the requirements on the over-all functionality.
The COCONUT project is an example of such software development project in the domain of numerical optimization and constraint programming.

The authors of \cite{coop:granvilliers-monfroy-benhamou-issac-01} 
illustrate practical importance of \mbox{analysis} of composite solvers with examples from interval constraint programming.
They point out two aspects of this analysis:
(a) detection of good cooperation strategies subject to expectations from the composite solver and the individual solvers in hand;
(b) reasoning about the properties of individual solvers subject to the expectations from the composite solver.
In this paper we focus on the latter aspect.

Certain properties of composite solvers are expressible in terms of the properties of the processed {\em data}.
One can study the properties of this sort using the frameworks of software verification, programming logics, model checking program analysis.
In order to reason about the properties of {\em individual solvers} one needs a new kind of analysis.

In this paper we present a constraint-based \mbox{analysis} of the composite solvers
that integrates reasoning about the individual solvers and the processed data.
The idea is to approximate this reasoning by resolution of set constraints on the finite sets representing the predicates that express all the necessary properties.
We illustrate application of our analysis to two important cooperation patterns: deterministic choice and loop.

Before going further we give a quick motivational example.

\begin{example}[Motivation]

Suppose that we, rather ``engineers'' than experts in global optimization, 
have in hand a library of numerical algorithms that includes the standard methods for local and global optimization, different tests, etc.
and develop a software for minimization of quadratic functions (quadratic programming).
Algorithm \ref{motivation-ex} specifies the composite solver that we have built with the intention to avoid the costly exhaustive search in the case of convex objective functions.

\begin{algorithm}
\begin{algorithmic}
\IF{ $f$ is convex }
	\STATE solve $x=\arg\min_{x\ge 0} f$ by the method of steepest descent
\ELSE
	\STATE solve $x=\arg\min_{x\ge 0} f$ by the exhaustive search
\ENDIF
\end{algorithmic}
\caption
{	A na\"\i{}ve composite solver for quadratic programming.
}
\label{motivation-ex}
\end{algorithm}

Our logic is as follows:
if the graph of the objective is like a tea-cup (``convex''), then we use the steepest descent to go to its lowest point; otherwise we do the exhaustive search.
Will this work?
Well, sometimes: we have not thought of the objectives that are convex and unbounded from below in the orthant $x\ge 0$, e.g. $f(x_1,x_2)=(x_1-x_2)^2-x_1$.
The method of steepest descent may fail to terminate in this case.

How can we detect similar situations automatically?
How do we change a composite solver in order that it work correctly?
In this paper we propose a formalism suitable for this kind of \mbox{analisys}.

\end{example}

The paper is structured as follows.
In Section \ref{coop-model} we introduce the basic notions of the paper and describe the operation of composite solvers.
In Section \ref{infinite-calc} we describe a technique for reasoning about composite solvers in terms of constraints on finite sets.
In this Section \ref{examples} we explain how to express certain properties of the composite solvers in terms of set constraints.
Section \ref{conclusion} concludes the paper.

\section{Solvers, contexts, operation model}
\label{coop-model}

In this section we introduce the basic notions of the paper and describe the operation of composite solvers.

A set of individual solvers that interact and exchange some data through a shared data store is called {\em composite solver}.
We number the individual solvers by integers from 1 to $n$.
The content of the data store, called {\em context}, is divided into {\em control data} and {\em application specific data}.
The control data specify the set of the individual solvers that ``are called''.
In the constraint programming framework the application specific data are, usually, specifications in some declarative language.
The set of all contexts is denoted $\C$.

The operation of the composite solver is divided into ticks.
At the beginning of a tick every solver $s$ checks the control data and either changes some part of the context if it is called, or does nothing otherwise.
The initial context is provided by the user of the composite solver.
Thus our composite solvers are either sequential or synchronous bulk parallel systems.

A solver $s$ determines a transformation $F_s:\C\to \C$ of the contexts at the beginning of the ticks into the contexts at the end of the ticks.
The synchronous modifications of the data store must be coherent, that is, $F_s(c)=F_{s'}(c)$
for any context $c$ that contains the control data indicating that solvers $s$ and $s'$ are called.

A context is {\em feasible}, if it is generated from the initial context by some sequence of the transformations $F_s$'s where $s$'s are some solvers.

Now we proceed to the description of our constraint-based formalism.

\section{From composite solvers to finite sets}
\label{infinite-calc}

In this section we describe a technique for reasoning about composite solvers in terms of constraints on finite sets.
We axiomatize the operation model from Section \ref{coop-model} in the first order logic and
view the axioms as set constraints on the interpretation of the symbols involved therein.
Since the exact solutions to these constraints may be (non-constructible explicitly) infinite sets,
we solve our set constraints approximately modulo a finite set of clusters of the contexts $\C$.
We assume that the reader is familiar with the generic concepts of constraint and constraint satisfaction.

The axioms in question are written in terms of
a constant symbol $c_0$ denoting the initial context,
a unary predicate symbol $p$ denoting the set of feasible contexts and
unary function symbols $f_s$'s denoting the transformations $F_s$.
The axioms are plain and open ($n$ is the number of individual solvers):
\begin{eqnarray}
&&p(c_0)															\label{first-ax}\\
&&\forall c\; p(c)\wedge f_s(c)=c' \Longrightarrow p(c') \quad s=1,\ldots,n							\label{next-ax}\\
&&\forall c' \exists c\; p(c') \wedge c'\ne c_0 \Longrightarrow \left(c'=f_1(c)\vee\ldots\vee c'=f_n(c)\right)\wedge p(c)	\label{last-ax}
\end{eqnarray}

The axiom (\ref{first-ax}) states that the initial context is feasible.
The $n$ axioms (\ref{next-ax}) say that the transformations $F_s$'s map feasible contexts onto themselves.
The axiom (\ref{last-ax}) states that every feasible context except the initial one has a feasible pre-image under some transformation $F_s$.

The axioms (\ref{first-ax})--(\ref{last-ax}) are nothing else than 
constraints on the interpretation $\dot{c_0}$, $\dot{p}$ and $\dot{f_s}$'s, in the model-theoretic sense, of the symbols $c_0$, $p$ and $f_s$'s.
One can write down these constraints are as follows:
\begin{eqnarray}
&\fun\left(\dot{f_1}\right), \ldots, \fun\left(\dot{f_n}\right)&								\label{first-cnt}\\
&\dot{p} = \{\dot{c_0}\} \cup \img\left(\dot{f_1},\dot{p}\right) \cup \cdots \cup \img\left(\dot{f_n},\dot{p}\right)&		\label{last-cnt}
\end{eqnarray}
The domains of $\dot{c_0}$, $\dot{p}$ and $\dot{f_s}$'s consist of the contexts $\C$, of all the subsets of $\C$ and, respectively, of all the binary relations on $\C$.
The symbols $=$ and $\cup$ denote equality and union of subsets of $\C$.
The symbol $\fun$ denotes the constraint ``is a function from $\C$ to $\C$''.
The symbol $\img$ denotes the image of a subset of $\C$ under a binary relation on $\C$.

Since many exact solutions to the constraints (\ref{first-cnt})--(\ref{last-cnt}) involve infinite sets,
we sacrifice precision for tractability and group the individual contexts from $\C$ into finitely many clusters called {\em context properties}, denoted $\C^\star$.
A binary relation on $\C^\star$ whose domain is $\C^\star$ is called {\em abstract solver}.
A context property $c^\star$ {\em approximates} a context $c$, iff $c\in c^\star$.
A set $P^\star$ of context properties {\em approximates} a set $P\subseteq\C$, iff $P\subseteq \cup P^\star$.
An abstract solver $R^\star$ {\em approximates} a binary relation $R\subseteq \C^2$, iff $R\subseteq \cup R^\star$.

\begin{example}[Hull Consistency in sharpness analysis]

\def\acycl{{\tt tree}}
\def\compl{{\tt ok}}

Consider the Hull Consistency (HC) algorithm \cite{opt:benhamou-goualard-granvilliers-iclp-99} from interval constraint programming.
Given a set $i$ of interval constraints, this algorithm computes a box $b$ that bounds the set $\sol(i)$ of the solutions to $i$.
Let the context specify the constraints $i$, the box $b$, and the necessary control data (of no interest at the moment).

The well-known fact about the HC algorithm is that it bounds the set $\sol(i)$ sharply, i.e. $b$ cannot be improved without losing a solution to $i$,
if $i$ has an acyclic constraint graph (see \cite{intervals:hansen-reliable-comp-97}).
We can express this fact in terms of the context properties ``$i$ has an acyclic constraint graph'' (abbreviated $\acycl$) and ``$b=\sol(i)$'' (abbreviated $\compl$)
by the abstract solver $HC^\star=\left\{\left(\acycl,\compl\right)\;\left(\C,\C\right)\;\left(\compl,\compl\right)\right\}$.
For example, $(\acycl,\compl) \in HC^\star$ means that the HC algorithm bounds $\sol(i)$ sharply, if $i$ has an acyclic constraint graph.

\end{example}

\begin{proposition}[Correctness]
If $\dot{c_0}\in\C^\star$, $\dot{p}\subseteq \C^\star$ and abstract solvers $\dot{f_s}$'s approximate some solution to the constraints (\ref{first-cnt})--(\ref{last-cnt}),
then they satisfy these same constraints with respect to the following definition of $=$, $\cup$, $\fun$, $\img$:
\begin{eqnarray*}
\fun(R^\star) & 	\iff		& \mbox{$R^\star$ is an abstract solver}\\
\img(R^\star,P^\star) & =		& \mbox{the image of} \; P^\star \; \mbox{under} \; R^\star\\
P^\star \cup Q^\star &	=		& \mbox{the standard union of subsets of $\C^\star$}\\
P^\star = Q^\star & \iff		& \mbox{the standard equality of subsets of $\C^\star$}
\end{eqnarray*}

\end{proposition}

In practice the constraints (\ref{first-cnt})--(\ref{last-cnt}) are joined to (some of) the constraints
\begin{eqnarray}
&\dot{f_1}=F_1^\star, \ldots, \dot{f_n}=F_n^\star&								\label{optional-cnt}
\end{eqnarray}
specifying the abstract solvers.
We build these latter $F_s^\star$'s using two data bases
that contain patterns of the individual solvers and the relation of logical equivalence on the set of properties of the processed data.

A {\em pattern} of an individual solver is a collection of {\em rules} of the form ``pre-condition $\to$ post-condition''
that have as common formal parameters the processed data and the called solvers.
The pre- and post-condition are conjunctions of atomic formulas containing the formal parameters of the pattern.
The formal parameters corresponding to the data not modified by the solver can be marked as ``read-only''.

The fact that some solver is called is expressed by the unary predicate symbol $\called$; the interpretation of other predicate symbols is arbitrary.
The pre- and post-conditions in the pattern of a solver $s$ are of the form $\called(s)\wedge C$ and the symbol $\called$ does not occur in the conjunction $C$.
Thus parallelism is not actually allowed.

\def\rhs{{\tt rhs}}

Let $\pi_1$, \ldots, $\pi_n$ be the patterns of the individual solvers with instantiated formal parameters.
The context properties are conjunctions $\called(s)\wedge C$,
 ---where $s=1, \ldots, n$ and $C$ is a conjunction of the atomic formulas from $\pi_1$, \ldots, $\pi_n$,---
that are not equivalent to the false conjunction.
We assume that two equivalent conjunctions are the same object.

The image of a context property $c^\star$ under the abstract solver $F_s^\star$ corresponing to a pattern $\pi_s$ is built as follows.
Let $c^\star = \called(s') \wedge C_{\tt ro} \wedge C_{\tt rw}$ such that
every conjunct in $C_{\tt rw}$ contains a value taken by some non read-only formal parameter of $\pi_s$ and
$C_{\tt ro}$ contains all the other conjuncts from $c^\star$ except $\called(s')$.

If $s\ne s'$ then $\img\left( F_s^\star, \left\{ c^\star \right\} \right) = \left\{ c^\star \right\} $.
Otherwise $\img\left( F_s^\star, \left\{ c^\star \right\} \right)$ =
$\bigwedge \left\{ \left\{ C_{\tt ro} \right\} \right\} \cup \left\{ \rhs(c^\star_1, \pi_s) | c^\star_1 \; \mbox{is implied by} \; c^\star \right\}$
where each set of conjunctions is interpreted as the disjunction of its elements,
e.g. $\{a,b\}\wedge \{x,y\}=\{a \wedge x, a \wedge y, b \wedge x, b \wedge y\}$,
and $\rhs(c^\star_1, \pi_s)$ denotes the set of the post-conditions following a pre-condition $c^\star_1$ in the pattern $\pi_s$.
Finally, $c^\star_1$ ``is implied by'' $c^\star$ iff $c^\star_1 \wedge c^\star$ is equivalent to $c^\star$.

The next section illustrates our approach by several examples.

\section{Examples}
\label{examples}

The examples in this section illustrate application of our approach to two important cooperation patterns: deterministic choice and loop.
Our ultimate goal (out of the scope of this paper) is to couple the analysis with the language for specification of composite solvers in the framework of the COCONUT project.
In order that the reader can feel our approach better, we provide in Appendix \ref{logicalc} the complete specification of the set constraints from Section \ref{cplex-ac}.

\subsection{The na\"\i{}ve solver from Section \ref{intro}}

The patterns of the individual solvers from the example in Section \ref{intro} are as follows:
{
\small
\begin{eqnarray*}
&&{\tt cnvx?}( {\tt ro}(F); S_1, S_2 ) =
		\{
			\called(1) \wedge {\tt cnvx}(F)		\to \called(S_1),
			\called(1)				\to \called(S_2)
		\}\\
&&{\tt dscnt}( {\tt ro}(F), X; S ) =
		\{
			\called(2) \wedge {\tt stCnvx}(F)	\to \called(S) \wedge {\tt min}(F, X),
			\called(2)				\to \called(S)
		\}\\
&&{\tt glblSrch}( {\tt ro}(F), X; S ) =
		\{
			\called(3)				\to \called(S) \wedge {\tt min}(F, X)
		\}\\
&&{\tt done}() =
		\{
			\called(4)				\to \called(4)
		\}
\end{eqnarray*}
}
The symbols ${\tt cnvx}$, ${\tt stCnvx}$, ${\tt min}$ denote the properties ``is convex'', ``is strictly convex'', ``is the global minimizer in the positive orthant''.
The only non-trivial equivalence is ${\tt cnvx}(F)\wedge {\tt stCnvx}(F)\equiv {\tt stCnvx}(F)$.
The read-only parameters are marked ${\tt ro}$.

The instantiated patterns are ${\tt cnvx?}( f; 2, 3 )$, ${\tt dscnt}( f, x; 4 )$, ${\tt glblSrch}( f, x; 4 )$, ${\tt done}()$.
The set of context properties is (we use the notation for disjunctions from Section \ref{infinite-calc}):
\{$\called(1)$, $\called(2)$, $\called(3)$, $\called(4)$\}
$\wedge$
\{${\tt cnvx}(f)$, ${\tt stCnvx}(f)$, ${\tt min}(f,x)$, ${\tt true}$\}
$\wedge$
\{${\tt min}(f,x)$, ${\tt true}$\}.
In practice these 24 context properties are numbered and the set constraints involve only their numbers.

Of the specifications for $F_1^\star$, $F_2^\star$, $F_3^\star$, $F_4^\star$ generated by the procedure from Section \ref{infinite-calc}, we provide the first one:
{\small
\begin{eqnarray*}
&&\img( F_1^\star, \{ \called(1) \wedge {\tt cnvx}(f) \} ) = \{ \called(2) \wedge {\tt cnvx}(f)\}\\
&&\img( F_1^\star, \{ \called(1) \wedge {\tt stCnvx}(f) \} ) = \{ \called(2) \wedge {\tt stCnvx}(f)\}\\
&&\img( F_1^\star, \{ \called(1) \wedge {\tt min}(f,x) \wedge {\tt cnvx}(f) \} ) = \{ \called(2) \wedge {\tt min}(f,x) \wedge {\tt cnvx}(f) \}\\
&&\img( F_1^\star, \{ \called(1) \wedge {\tt min}(f,x) \wedge {\tt stCnvx}(f)\} ) = \{ \called(2) \wedge {\tt min}(f,x) \wedge {\tt stCnvx}(f) \}\\
&&\img( F_1^\star, \{ \called(1) \wedge {\tt min}(f,x) \} ) = \{ \called(2) \wedge {\tt min}(f,x) \wedge {\tt cnvx}(f), \called(3) \wedge {\tt min}(f,x) \}\\
&&\img( F_1^\star, \{ \called(1) \} ) = \{ \called(2) \wedge {\tt cnvx}(f), \called(3) \}
\end{eqnarray*}}
and $\img( F_1^\star, \{ c^\star \} ) = \{ c^\star \}$ for the other context properties $c^\star$.

Solving the constraints (\ref{first-cnt})--(\ref{optional-cnt}) and $\dot{c_0}=\called(1)$,
we obtain the following approximation for the set of feasible contexts:
$\dot{p}$ = \{$\called(1)$, $\called(2)$ $\wedge$ ${\tt cnvx}(f)$, $\called(3)$, $\called(4)$, $\called(4)$ $\wedge$ ${\tt min}(f,x)$\}.
Since this $\dot{p}$ contains $\called(4)$, we are not sure that our composite solver always finds the minimizer of $f(x)$ subject to $x\ge 0$.

The question ``When does our composite finds the minimizer?'' is translated into the constraints (\ref{first-cnt})--(\ref{optional-cnt}) and 
the constraints that 
say that we examine what happens after the convexity test is called ($\called(1)$ ``is implied by'' $\dot{c_0}$)
and
forbid the uncertain situation after termination ($\called(4) \not \in \dot{p}$).
Solving these constraints for the initial context $\dot{c_0}$, we obtain the following solutions: 
$\dot{c_0}=\called(1) \wedge {\tt stCnvx}(f)$,
$\dot{c_0}=\called(1) \wedge {\tt min}(f,x) \wedge {\tt stCnvx}(f)$.
This means that the objective has to be strictly convex in order that our solver can find its global minimizer.

\subsection{The Simplex method and Hull Consistency}
\label{cplex-ac}

Consider a composite solver that makes cooperate the Simplex method from linear programming and the HC algorithm \cite{opt:benhamou-goualard-granvilliers-iclp-99}
(a similar composite solver is described e.g. in \cite{coop:beringer-backer-lp-elsevier-95}).
The context specifies some linear, interval and bound constraints, denoted $\ell$, $i$ and, respectively $b$.
The Simplex method updates $b$ by bounding the solution set $\sol(\ell \cup b)$ and calls the HC algorithm,
which in its turn updates $b$ by bounding the solution set $\sol(i\cup b)$ and calls the Simplex method, and so on until stabilization of $b$.
The important quality of this strategy is that bounds the solution set $\sol(l\cup i\cup b)$ more sharply than the HC algorithm.

The patterns of the individual solvers are as follows:
{\small
\begin{eqnarray*}
&&{\tt cplex}( {\tt ro}(L), B; S ) =
	\left\{
		\called(1)				\to \called(S) \wedge {\tt ok}(L)
	\right\}\\
&&{\tt hc}( {\tt ro}(I), B; S ) =
	\left\{
		\called(2)				\to \called(S),
		\called(2) \wedge {\tt tree}(I)		\to \called(S) \wedge {\tt ok}(I)
	\right\}\\
&&{\tt same?}( {\tt ro}(B); S_1, S_2 ) =
	\left\{
		\called(3)				\to \called(S_1),
		\called(3)				\to \called(S_2)
	\right\}\\
&&{\tt done}() =
	\left\{
		\called(4)				\to \called(4)
	\right\}
\end{eqnarray*}}
The symbols ${\tt ok}$, ${\tt tree}$ denote the properties ``has the solution set that we can bound sharply'', ``has an acyclic constraint graph''.
All the equivalences are trivial.

The instantiated patterns are ${\tt cplex}( \ell, b; 2 )$, ${\tt hc}( i, b; 3 )$, ${\tt same?}( b; 1, 4 )$, ${\tt done}()$.
There are 32 context properties built as follows:
\{$\called(1)$, $\called(2)$, $\called(3)$, $\called(4)$\}
$\wedge$
\{${\tt ok}(\ell)$, ${\tt ok}(i)$, ${\tt tree}(i)$, ${\tt true}$\}
$\wedge$
\{${\tt ok}(i)$, ${\tt tree}(i)$, ${\tt true}$\}
$\wedge$
\{${\tt tree}(i)$, ${\tt true}$\}.

The specifications for the abstract solvers $F_1^\star$, $F_2^\star$, $F_3^\star$, $F_4^\star$ generated by the procedure from Section \ref{infinite-calc}
are provided in Appendix \ref{logicalc}.

Solving the constraints (\ref{first-cnt})--(\ref{optional-cnt}) and $\dot{c_0}=\called(1)$,
we obtain the following approximation for the set of feasible contexts:
$\dot{p}=\{ \called(1), \called(1) \wedge {\tt ok}(\ell), \called(2) \wedge {\tt ok}(\ell), \called(3) \wedge {\tt ok}(\ell), \called(4) \wedge {\tt ok}(\ell)\}$.
Since this $\dot{p}$ contains only $\called(4) \wedge {\tt ok}(\ell)$,
the solution set of the linear constraints is {\em always} bounded sharply after termination of the composite solver.

We can figure out when the composite solver bounds the solution set $\sol(\ell\cup i\cup b)$ sharply, solving the constraints (\ref{first-cnt})--(\ref{optional-cnt}) and
the constraints that say
that we examine what happens after the Simplex method is called ($\called(1)$ ``is implied by'' $\dot{c_0}$) and
that the solution set is bounded sharply after termination
($\dot{c}_\infty \in \dot{p}$, $\called(4) \wedge {\tt ok}(i) \wedge {\tt ok}(\ell)$ ``is implied by'' $\dot{c}_\infty$).
These constraints have 6 solutions.
The first two assign $\called(4)$ $\wedge$ ${\tt ok}(\ell)$ $\wedge$ ${\tt ok}(i)$ to $\dot{c}_\infty$ and
either $\called(1)$ $\wedge$ ${\tt ok}(i)$, or $\called(1)$ $\wedge$ ${\tt ok}(i) \wedge {\tt ok}(\ell)$ to $\dot{c_0}$.
The other four assign $\called(4)$ $\wedge$ ${\tt ok}(\ell)$ $\wedge$ ${\tt tree}(i)$ $\wedge$ ${\tt ok}(i)$ to $\dot{c}_\infty$ and
one of the 4 context properties that ``imply'' $\called(1)$ $\wedge$ ${\tt tree}(i)$ to $\dot{c_0}$.
This means that,
the composite solver bounds the solution set $\sol(\ell\cup i\cup b)$ sharply, if the interval constraints $i$ have an acyclic constraint graph.

\section{Conclusion}
\label{conclusion}

We have presented a formalism for automatic analysis of composite solvers.
This formalism provides
a structure for expressing properties of the data store (context properties),
a structure for specifying the behaviour of solvers (abstract solvers),
a method for approximation of composite solvers by set constraints
that can be efficiently solved by conventional set constraint solvers like \cite{Gervet97,petrov-yakhno-lncs-1755-99}.
The ultimate goal (out of the scope of this paper) is to couple our analysis with the language for specification of composite solvers in the framework of the COCONUT project.

\paragraph*{Acknowledgements}
We thank Lucas Bordeaux from l'Institut de Recherche en Informatique de Nantes for the comments on quantified constraints.

\paragraph*{Disclaimer}
No solver has been damaged during the preparation of this document.

\bibliographystyle{abbrv}

\appendix

\section{Specification of the example from Section \ref{cplex-ac}}
\label{logicalc}

The constraints from the sharpness example are provided in Fig. \ref{logicalc-spec} in the LogiCalc language \cite{petrov-yakhno-lncs-1755-99}.
We recall its syntax/semantics.
The LogiCalc language allows the user to specify constraints on integer numbers, tuples and finite sets.
Tuples of sets, sets of tuples, sets of sets, etc. are allowed.
The constraints are specified in terms of set inclusion ${\tt subset}$, membership ${\tt in}$, equality ${\tt =}$, and inequality ${\tt <=}$.
The left and right hand sides of the constraints are expressions built from variables, arithmetic and set operations, and specifications of finite set.

Finite sets are specified by either enumeration of the elements, or by their common property.
For example,
{\small
\begin{verbatim}
x subset { 2, 3, 5, 7 };
y = { i * j | i in x; j in x; i + 1 <= j };
y = { 6, 10, 15 }
\end{verbatim}}
specify the set $y=\{ i\cdot j | i\in x, j\in x, i < j \}=\{6,10,15\}$ and the set $x=\{2,3,5\}$ of their prime factors.
Notice implicit existential quantification of the variables $i$ and $j$ that do not occur outside the specification of $y$.
Notice that the equation $y=\{ i\cdot j | \ldots \}$ is, in fact, a constraint on $x$ and $y$.

\begin{figure}[ht]
{\small
\begin{verbatim}
% Notation for the data properties: %
treeI  = 0; okL  = 1; okI      = 2; treeIokL    = 3;
okIokL = 4; true = 5; treeIokI = 6; treeIokLokI = 7;
Cdata = { treeI, okL, okI, treeIokL,
          okIokL, true, treeIokI, treeIokLokI };
% cplex %
F1star = {
        ((1, treeI),       (2, treeIokL)),
        ((1, okL),         (2, okL)),
        ((1, okI),         (2, okIokL)),
        ((1, treeIokL),    (2, treeIokL)), 
        ((1, okIokL),      (2, okIokL)), 
        ((1, true),        (2, okL)),
        ((1, treeIokI),    (2, treeIokLokI))
        ((1, treeIokLokI), (2, treeIokLokI)) } \/ 
{ ((i1, z1), (i1, z1)) | i1 in { 2, 3, 4 }, z1 in Cdata };
% hc %
F2star = {
        ((2, treeI),       (3, treeIokI)),
        ((2, okL),         (3, okL)),
        ((2, okI),         (3, okI)),
        ((2, treeIokL),    (3, treeIokLokI)), 
        ((2, okIokL),      (3, okIokL)), 
        ((2, true),        (3, true)),
        ((2, treeIokI),    (3, treeIokI))
        ((2, treeIokLokI), (3, treeIokLokI)) } \/
{ ((i2, z2), (i2, z2)) | i2 in { 1, 3, 4 }, z2 in Cdata };
% same? %
F3star = { ((3, z3), (i3, z3)) | z3 in Cdata, i3 in { 1, 4 } };
% done %
F4star = { ((4, z4), (4, z4)) | z4 in Cdata };
% constraints (4)--(5): %
img1 = { c11 | (c1, c11) in F1star; c1 in p };
img2 = { c22 | (c2, c22) in F2star; c2 in p };
img3 = { c33 | (c3, c33) in F3star; c3 in p };
img4 = { c44 | (c4, c44) in F4star; c4 in p };
p = { c0 } \/ img1 \/ img2 \/ img3 \/ img4;
\end{verbatim}
}
\caption{
	Specification of the sharpness example in the LogiCalc language;
	${\tt\{\}}$ denotes the empty set,
	${\tt \symbol{'134}/}$ denotes set union,
	${\tt F1star}$ specifies the abstract CPLEX,
	${\tt F2star}$ specifies the abstract arc consistency,
	${\tt c0}$ denotes the initial context that we search for,
	${\tt img1}$, ${\tt img2}$ denote the images of the set of feasible contexts under the abstract CPLEX and arc consistency.}
\label{logicalc-spec}
\end{figure}

\end{document}